\documentclass{article}
\usepackage{enumerate}
\usepackage{microtype}
\usepackage{graphicx}
\usepackage{booktabs} 
\usepackage{makecell}
\usepackage{graphicx}
\usepackage{caption}
\usepackage{courier}
\usepackage{multirow}
\usepackage[usenames]{color}
\usepackage{fullpage}
\usepackage{subcaption}
\usepackage{todonotes}
\usepackage{amssymb}
\usepackage{gensymb}
\usepackage{amsfonts}       
\usepackage{enumitem}
\usepackage{multirow}
\usepackage{algorithmic}
\usepackage[linesnumbered, ruled]{algorithm2e}
\usepackage{amsmath,nccmath}
\usepackage[colorlinks,linkcolor=red,anchorcolor=blue,citecolor=blue]{hyperref}

\begin{document}
\title{Transferable Neural Processes for Hyperparameter Optimization}
\author
{
Ying Wei\thanks{Tencent AI Lab, Shenzhen, China; e-mail: {\tt judyweiying@gmail.com}}
		~,~
	Huaxiu Yao\thanks{College of Information Science and Technology, Pennsylvania State University, State College, PA 16801, USA; e-mail: {\tt huaxiuyao@psu.edu}} 
	~,~
	Peilin Zhao\thanks{Tencent AI Lab, Shenzhen, China; e-mail: {\tt masonzhao@tencent.com}}
		~,~
	Junzhou Huang\thanks{Tencent AI Lab, Shenzhen, China; e-mail: {\tt jzhuang@uta.edu}}
}
\date{}
\maketitle

\begin{abstract}
Automated machine learning 
aims
to automate the whole process of machine learning, including 
model configuration. 
In this paper, we focus on automated hyperparameter optimization (HPO) based on sequential model-based optimization (SMBO).
Though conventional SMBO algorithms
work well when abundant 
HPO
trials are available, they are far from satisfactory in practical applications where a trial on a huge dataset may be so costly that an optimal hyperparameter configuration is expected to return in as few trials as possible.
Observing that human experts draw on their expertise in a machine learning model by trying configurations that once performed well on other datasets, 
we are inspired to speed up 
HPO
by transferring knowledge from 
historical 
HPO trials on other datasets.
We propose an end-to-end and efficient 
HPO algorithm named as Transfer Neural Processes (TNP), 
which 
achieves 
transfer learning by incorporating 
trials on other datasets, initializing the model with well-generalized parameters, and learning an initial set of hyperparameters to evaluate. 
Experiments on extensive OpenML datasets and three computer vision datasets show that the proposed model can achieve state-of-the-art 
performance 
in at least one order of magnitude less trials.

\end{abstract}

\section{Introduction}
In the pipeline of a machine learning system, model configuration poses daunting challenges: 1) how to choose the optimal algorithm among hundreds to thousands of machine learning algorithms?
2) how to configure the optimal hyperparameters after an algorithm is specified?
Brute-force exploration of all possible solutions, obviously, is prohibitively expensive and impractical.
Though experts can draw on their expertise to pinpoint a configuration relatively quickly, practitioners outside machine learning can get bogged down in the meticulous design.
These challenges highlight the critical importance of automating model configuration where we focus on automated hyperparameter optimization (HPO).

Researchers have explored
three 
strands 
of 
HPO methods including exhaustive search~\cite{bergstra2012random}, model-specific methods~\cite{keerthi2007efficient} and sequential model-based optimization (SMBO)~\cite{jones1998efficient}.
Free from the limitations of exhaustive search being computationally expensive and model-specific methods being too customizable to be applied in general, 
SMBO 
has been the current state-of-the-art. 
The core of SMBO is to learn from observed hyperparameter performances a surrogate model which maps a hyperparamter configuration to the evaluation measure on a dataset.
Sequentially, in each trial, a promising 
configuration estimated by the surrogate is evaluated and this new 
observation is incorporated to further improve the surrogate.
While existing surrogate models including Gaussian Processes (GPs)~\cite{snoek2012practical}, parzen estimators~\cite{bergstra2011algorithms}, random forest~\cite{hutter2011sequential}, and neural networks~\cite{snoek2015scalable,springenberg2016bayesian}  have 
shown their effectiveness provided with 
sufficient 
observations, it is 
imperative to return an 
optimal 
configuration 
in very few 
trials, e.g., ten, in real-world applications 
where 
a trial on huge 
datasets is quite costly.

Inspired by the fact that human experts hone their skills in a machine learning model via training it 
on various datasets and apply the skills to the dataset at hand, 
we are devoted to another important  research direction to speed up the process of HPO, 
i.e., transferring knowledge from historical trials on other datasets.
The motivation behind 
is that a subset of hyperparameter configurations that perform well on some datasets, especially those bearing a striking similarity with the target dataset of interest, are likely qualified candidates for the target.  
As the convention of transfer learning~\cite{pan2009survey}, there exist three research problems, i.e., when, what, and how to transfer.
First, when to transfer here is grounded on measuring the similarity between datasets, so that negative transfer is avoided without leveraging knowledge from wildly dissimilar ones. 
Most of existing studies~\cite{bardenet2013collaborative,feurer2015initializing,schilling2015hyperparameter,yogatama2014efficient} rely on meta-features of a dataset which, however, are hand-crafted and 
loosely 
related to 
hyperparameter performances.
Second, despite the instantiation of what to transfer as  either initializations~\cite{feurer2015initializing,lindauer2018warmstarting,wistuba2015learning}, observations~\cite{schilling2015hyperparameter,swersky2013multi,yogatama2014efficient}, parameters of a surrogate model~\cite{bardenet2013collaborative,feurer2018scalable,perrone2018scalable,wistuba2016two}, or acquisition functions~\cite{wistuba2016hyperparameter}, none of  existing works is qualified to harness the collective power of them.
Third, in terms of how to transfer, almost all previous works develop GP-based surrogate models with cubic scaling which are highly inefficient and even impractical  to\textbf{} incorporate abundant past observations.
One rencet work~\cite{perrone2018scalable} applies Bayesian linear regression as GP approximation, but meanwhile it loses predictive power.
Especially, they all require an explicitly defined kernel, e.g., the linear kernel in~\cite{perrone2018scalable} which is fixed 
across datasets, 
being inadequate
to accomodate a heterogeneous dataset in practical scenarios.

To address these problems, we propose a novel end-to-end hyperparameter optimization algorithm called Transfer Neural Processes (TNP). 
Motivated by recent success of Neural Processes (NPs)~\cite{garnelo2018conditional}, we adopt NPs as our surrogate model.
By combining the best of both GP and neural networks, NPs can preserve the property reminiscent of GP, i.e., defining distributions over functions, and meanwhile be efficiently trained with standard deep learning libraries.
The TNP consists of an encoder which learns a representation of each observation, a dataset-aware attention unit which attentively aggregates representations of all observations to infer the latent distribution of hyperparameter performances, 
and a decoder which predicts the performances for target hyperparameter 
configurations 
with uncertainties by taking the latent distribution 
as input.
The proposed model achieves transfer learning by leveraging 
observations of 
previous datasets, learning from 
all datasets a transferable initialization for parameters of the TNP, as well as optimizing a well-generalized initial set of configurations to evaluate for SMBO.
The dataset-aware attention unit evaluates the
similarity between datasets 
using
the representations of all observations in a dataset 
and 
eliminates 
the need of manually defining meta-features.
Moreover, the parameters of the 
TNP modelling an implicit kernel 
are 
fine-tuned with several gradient updates for a target dataset, which empowers TNP to meet more wildly heterogeneous datasets.

\section{Related Work}
One
influential line of research to accelerate HPO 
is to leverage knowledge from  historical 
trials on other datasets that are similar to the target dataset of interest.
To measure the similarity between the target 
and previous datasets, the majority resort to manually defined meta-features of a dataset.
Feurer et al.~\cite{feurer2015initializing} propose to initialize a hyperparameter search with the best configurations from similar datasets.
Similarly, observations from 
$k$ nearest neighbour datasets in the meta-feature space are incorporated to train the surrogate model together with those in the target~\cite{schilling2015hyperparameter,yogatama2014efficient}.
Assuming a globally shared GP model, Bardenet et al.~\cite{bardenet2013collaborative} optimize the model with observations from all datasets.
Each observation is 
described as the concatenation of hyperparameters and meta-features.
The downside of these methods comes with the challenge of meta-features, i.e., being hand-crafted and loosely correlated to hyperparameter search behaviors.

There have been several attempts towards eliminating meta-features. For example, Wistuba et al.~\cite{wistuba2015learning} adopt 
a meta-loss to learn a set of 
initial configurations 
from past observations
to maximize the performance at the very beginning of SMBO.
In~\cite{swersky2013multi}, a multitask GP borrows observations of similar datasets where the similarity as a kernel is learned.
Besides, multiple GP experts each of which is trained on a previous dataset are combined to be the surrogate for the target dataset, where the ensemble weight 
is
learned as the generalization error of each expert on the target 
~\cite{feurer2018scalable,wistuba2016two}.
A more recent work~\cite{perrone2018scalable} conducts Bayesian linear regressions with a feature map learned by a neural network. The shared feature map is believed to improve knowledge generalization across datasets.
Unfortunately, all these works require a kernel to be explicitly defined, which gives rise to either poor scaling for GP-based approaches~\cite{feurer2018scalable,swersky2013multi, wistuba2015learning,wistuba2016two} or unfeasible algorithm deployment using standard deep learning libraries for linear kernel~\cite{perrone2018scalable}.
What is more, unlike ours, these works 
assuming 
the kernel as prior to be 
globally shared across datasets fail to accommodate heterogeneous datasets.

\section{Background and Problem Setup}
Given a dataset $\mathcal{D}\!\sim\! \mathcal{P}_{D}$, 
\emph{hyperparameter optimization (HPO)} 
aims
to identify optimal values
for hyperparameters $\mathbf{x}$ so that the generalization 
metric  
is maximized (e.g., accuracy) or minimized, i.e.,
\begin{equation}
\mathbf{x}^{*} = \arg\max_{\mathbf{x}\in\mathcal{X}} \mathbb{E}_{\mathbf{d}\sim\mathcal{P}_D}[\mathcal{L}(\mathbf{d},A_{\mathbf{x}}(\mathcal{D}))] =\arg\max_{\mathbf{x}\in\mathcal{X} } f(\mathbf{x}),
\label{eqn:hyperopt_1}
\end{equation}
where $\mathbf{d}$ is a sample drawn from $\mathcal{P}_D$, and $A_{\mathbf{x}}(\mathcal{D})$ represents the model produced by training an algorithm $A$ equipped with hyperparameters $\mathbf{x}$ on the dataset $\mathcal{D}$.
The hyperparameter space $\mathcal{X}$ could be continuous or discrete. 
Considering the difficulty of 
evaluating
the expectation over an unknown distribution $\mathcal{P}_D$ 
and optimizing it, 
a \emph{hyperparameter response function} $f$ $w.r.t.$ the hyperparameters $\mathbf{x}$ is maximized instead. 
HPO, in this case, is equivalent to maximizing the black-box function $f$ over $\mathcal{X}$, as there is no
knowledge of 
the response function $f$ and the search space $\mathcal{X}$.

\emph{Sequential Model-based Bayesian Optimizaion (SMBO)}~\cite{jones1998efficient} has been a dominant framework for global optimization of black-box functions.
SMBO consists of two components, i.e., a surrogate model $\Phi$ to approximate the response function and an acquisition function $a$ to determine the next hyperparameter configuration to evaluate.
Provided with $n_I$ initial configurations $\mathbf{x}_{I0},\!\cdots\!,\mathbf{x}_{In_I}$, SMBO starts by querying the values of the function $f$ at these configurations to constitute the initial set of history observations $\mathcal{H}_0\!=\!\{(\mathbf{x}_{I0}, y_{I0}),\!\cdots\!,(\mathbf{x}_{In_I}, y_{In_I})\}$. 
Afterwards, it iterates the following four stages:
1) in the $t$-th iteration (trial), 
fit the surrogate $\Phi_t$ on the observations $\mathcal{H}_{t-1}$;
2) 
use 
the surrogate $\Phi_t$ to make predictions $\{\hat{{\mu}}_{j}\}_{j=0}^{n_\mathcal{X}}$ with uncertainties $\{\hat{{\sigma}}_{j}\}_{j=0}^{n_\mathcal{X}}$ for 
$n_\mathcal{X}$ target configurations $\{\hat{\mathbf{x}}_{j}\}_{j=0}^{n_\mathcal{X}}$;
3) based on the predictions and uncertainties,
the acquisition function $a$ 
decides the next configuration $\mathbf{x}_{t}\!\in\!\{\hat{\mathbf{x}}_{j}\}_{j=0}^{n_\mathcal{X}}$ to try; 
4) evaluate the function $f$ at $\mathbf{x}_{t}$, and update the history set $\mathcal{H}_{t}\!=\!\mathcal{H}_{t-1}\!\cup\!\{(\mathbf{x}_{t},y_{t})\}$.


In the $t$-th iteration, 
there are a total number of $n_I + t$ observations in the history set $\mathcal{H}_{t-1}$.
In this paper, additionally, we leverage knowledge from $M$ history sets, i.e., $\mathcal{H}^1_{T^1},\!\cdots\!,\mathcal{H}^M_{T^M}$,  of HPO on $M$ datasets, i.e., $\mathcal{D}^1,\!\cdots\!,\mathcal{D}^M$.
In the $m$-th dataset $\mathcal{D}^m$, there are $T^m$ observations available in the history set  $\mathcal{H}^m_{T^m}=\{(\mathbf{x}^m_{t'},y^m_{t'})\}_{{t'}=0}^{T^m}$.
The goal of this paper lies that by borrowing strength from these $M$ history sets on $M$ 
datasets, 
the optimal hyperparameter configuration that maximizes 
the surrogate model 
(equivalent to maximizing the response function $f$) 
can be quickly returned in 
only a few 
trials.

\section{Transferable Neural Processes}
In this section, we detail the proposed
Transferable Neural Processes (TNP) 
by first introducing 
the neural process model as the surrogate $\Phi_t$ and illustrating how we innovatively fit the surrogate on a current observation set 
$\mathcal{H}_{t-1}$.
Next, we highlight how we accelerate maximizing the surrogate by simultaneously
taking
advantage of three types of knowledge transferred from historical HPO tasks on other datasets, i.e., observations, parameters for the surrogate, and the initial set of configurations. 
\subsection{The Neural Process Model}
\label{sec:npm}
The Neural Processes (NPs)~\cite{garnelo2018conditional,garnelo2018neural,kim2019attentive}, as a neural alternative to GPs, approaches regression by 
learning
a distribution over functions that map inputs to outputs instead of a single function.
As a result, the NPs provides uncertainty estimation 
besides a predicted response function value for a hyperparameter configuration.
Meanwhile, the NPs enjoys the desirable advantages of neural networks, including the efficiency in adapting to a newly incorporated observation 
and the linear scalability with regard to the number of observations.

\begin{figure*}[t]
	\centering
 	\includegraphics[width=\textwidth]{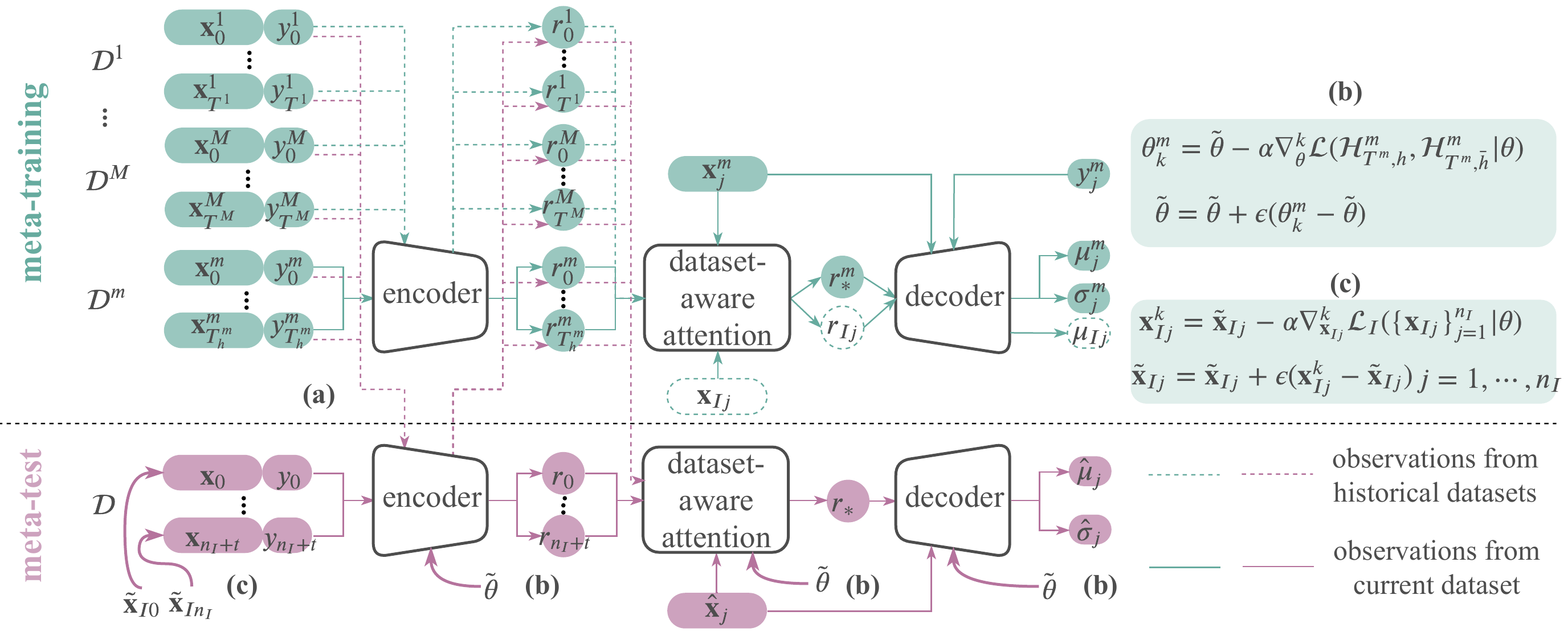}
\caption{The \textbf{Transferable Neural Processes}
consists of two stages. In the meta-training stage (colored as green), HPO trials on $M$ historical datasets are leveraged to \textbf{(b)} learn the transferable initializations for parameters of the 
TNP (i.e., $\tilde{\theta}$) and \textbf{(c)} optimize the well-generalized initial configurations for SMBO (i.e., $\{\tilde{\mathbf{x}}_{I0},\!\cdots\!,\tilde{\mathbf{x}}_{In_I}\}$ ).
During the meta-test stage (colored as purple), besides drawing on 
the initializations for TNP and the initial configurations for SMBO learned in  meta-training, the TNP also \textbf{(a)} takes all historical
observations from $M$ datasets into consideration. 
Remarkably, the TNP further fine-tunes the parameters initialized with $\tilde{\theta}$ by training on the current observation set $\mathcal{H}_{t-1}$, which allows the prior to be quickly tailored for the target dataset of interest. 
Best viewed in color.
}
	\label{fig:framework}
\end{figure*}

Motivated by 
NPs~\cite{garnelo2018conditional,garnelo2018neural,kim2019attentive},
we propose a neural process model involving 
three components, as shown  
in Figure~\ref{fig:framework}.
For the target dataset $\mathcal{D}$ of interest, 
the encoder 
learns an embedding $\mathbf{r}_{t'}\!\in\!\mathbb{R}^r$ for each observation $(\mathbf{x}_{t'},y_{t'})$, i.e., $\mathbf{r}_{t'}=E_{\theta_e}(\mathbf{x}_{t'},y_{t'})$, $\forall t'\!\in\!\{0,\!\cdots\!,n_I+t\}$.
Note that the encoder $E_{\theta_e}$ is parameterized with a neural network.
The dataset-aware attention unit as the second component summarizes all observations and produces an order-invariant representation of historical observations.
Mathematically, $\mathbf{r}_*=A_{\theta_a}(\mathbf{r}_0,\!\cdots\!,\mathbf{r}_{n_I+t})$.
This representation, $\mathbf{r}_*\!\in\mathbb{R}^r\!$, is expected to encode the latent distribution of hyperparameter performances conditioned on the 
set of observations $\mathcal{H}_{t-1}$.
We will detail this unit $A_{\theta_a}$ with an attention scheme later in Section~\ref{section:kt}.
Last but not the least, the decoder takes a target configuration  $\hat{\mathbf{x}}_{j}$  as well as  the representation $\mathbf{r}_*$ 
as input, 
and outputs the predicted value of the response function $f$ for this configuration, i.e., 
$\hat{\mathbf{y}}_{j}=D_{\theta_d}(\mathbf{r}_*, \hat{\mathbf{x}}_{j})$.
The prediction $\hat{\mathbf{y}}_{j}\!\in\!\mathbb{R}^2$
consists of 
two values 
which 
represent the mean $\hat{\mu}_{j}$ and variance $\hat{\sigma}_{j}$ of a Gaussian distribution $\mathcal{N}(\hat{\mu}_{j},\hat{\sigma}_{j})$, respectively.
We also parameterize the decoder $D_{\theta_d}$ with a neural network.

Draw inspiration from~\cite{garnelo2018neural}, we train the parameters of 
the neural process model, 
i.e., $\theta=\theta_e\cup\theta_a\cup\theta_d$, by following three steps:
1) randomly shuffle observations in $\mathcal{H}_{t-1}$ and divide them into two parts, e.g., $\mathcal{H}_{t-1,h}\!=\!\{(\mathbf{x}_{t'},y_{t'})\}_{t'=0}^{t_h}$ and $\mathcal{H}_{t-1,\bar{h}}\!=\!\{(\mathbf{x}_{t'},y_{t'})\}_{t'=t_h+1}^{n_I+t}$;
2) 
make predictions for  the configurations in 
$\mathcal{H}_{t-1,h}$, i.e., $\{\mathbf{x}_{t'}\}_{t'=0}^{t_h}$, conditioned 
on the observation set $\mathcal{H}_{t-1,\bar{h}}$;
3) maximize the conditional log likelihood,
\begin{align}
\label{eqn:loss}
\mathcal{L}(\mathcal{H}_{t-1,h},\mathcal{H}_{t-1,\bar{h}}|\theta) = 
\mathbb{E}_{f\sim P}[\mathbb{E}_{t_h}[\log{p_{\theta}(\{y_{t'}\}_{t'=0}^{t_h}|\mathcal{H}_{t-1,\bar{h}}, \{\mathbf{x}_{t'}\}_{t'=0}^{t_h})}]].
\end{align}
Practically, the gradient of the loss is 
estimated 
by sampling different response functions $f$ (equal to sampling datasets) 
and sampling different values of $t_h$. 

\subsection{Knowledge Transfer}
\label{section:kt}
\paragraph{Dataset-aware attention for leveraging observations}
The crux of GPs as the surrogate 
lies in modelling the similarity between a target configuration and the configurations of historical observations -- if a target configuration $\hat{\mathbf{x}}_{j}$ is close to the configuration of the $t'$-th observation $\mathbf{x}_{t'}$, its prediction $\hat{y}_j$ is expected to be close to $y_{t'}$.
Inspired by this, Kim et al.~\cite{kim2019attentive} modeled the similarity in NPs with the multihead attention mechanism~\cite{vaswani2017attention}.
Unfortunately, 
at the 
beginning of SMBO, say $t=1$, the number of observations in $\mathcal{H}_{t-1}$ is quite small, so that it is 
challenging to make accurate predictions. 
Therefore, we are motivated to also incorporate 
 abundant historical HPO observations on other datasets,
 namely that we consider both in-dataset observations and across-dataset observations.

To 
leverage across-dataset observations, 
we have to accommodate another desiderata: similarity 
between datasets.
Even if a target configuration $\hat{\mathbf{x}}_{j}$ is close to $\mathbf{x}^m_{t'}$, it is likely that the $t'$-th observation from the $m$-th dataset contributes little if the $m$-th and the target dataset are wildly different.
In fulfillment of this, we 
design our dataset-aware attention 
$A_{\theta_a}$ as,
\begin{align}
\mathbf{r}_{*}=\!A_{\theta_a}(\mathbf{r}_0,\!\cdots\!,\mathbf{r}_{n_I+t},\mathbf{r}_0^1,\!\cdots\!,\mathbf{r}^M_{T^M})
=\!\textbf{MultiHead}(g(\hat{\mathbf{x}}_{j}), g(\mathbf{X}^{0:M}),\mathbf{R}^{0:M},\mathbf{s}). \nonumber
\end{align}
$\mathbf{X}^{0:M}\!=\![\mathbf{X};\mathbf{X}^1;\!\cdots\!;\mathbf{X}^M]$ here includes the configurations of both in-dataset observations $\mathbf{X}\!=\!\{\mathbf{x}_{t'}\}_{t'=0}^{n_I+t}$ and across-dataset  ones $\mathbf{X}^m\!=\!\{\mathbf{x}^m_{t'}\}_{t'=0}^{T^m}$ ($\forall m\!=\!1,\!\cdots\!,M$).
Correspondingly, $\mathbf{R}^{0:M}\!=\![\mathbf{R};\mathbf{R}^1;\!\cdots\!;\mathbf{R}^M]$ with $\mathbf{R}\!=\!\{\mathbf{r}_{t'}\}_{t'=0}^{n_I+t}$ and $\mathbf{R}^m\!=\!\{\mathbf{r}^m_{t'}\}_{t'=0}^{T^m}$ ($\forall m\!=\!1,\!\cdots\!,M$) 
consists of the embeddings of all observations.

The multi-head attention $\textbf{MultiHead}$ is the concatenation of $H$ heads with each head $\text{head}_h\in\mathbb{R}^{d_h}$, where $d_h = r/H$.
Specifically, 
the $h$-th head 
follows $\text{head}_h :=\text{softmax}\big(\mathbf{s} \circ \big[g(\hat{\mathbf{x}}_{j})\mathbf{W}^q_h\big][g(\mathbf{X}^{0:M})\mathbf{W}^k_h]^T / \sqrt{r}\big)\mathbf{R}^{0:M}\mathbf{W}^v_h$,
where 
the similarity between in-dataset configurations is measured with scaled dot prodcut while that between across-dataset configurations is modulated by the similarity $\mathbf{s}$ between datasets.
Here 
$\mathbf{W}^q_h$, $\mathbf{W}^k_h$, $\mathbf{W}^v_h\in\mathbb{R}^{r\times d_h}$ are parameters;
$g(\cdot)$, called the key function and parameterized as an MLP, constructs a $r$-dimensional embedded representation of the configuration to empower a better comparison between the query configuration $\hat{\mathbf{x}_j}$ and the keys $\mathbf{X}^{0:M}$.

 We 
 especially highlight $\mathbf{s}$ which measures the similarity between the target and all datasets,
 i.e., $\mathbf{s}\!
 =\!\text{softmax}([1, s^1\mathbf{1}^{(1\times T^1)}, \!\cdots\!, s^M\mathbf{1}^{(1\times T^M)}])$
where $\mathbf{1}^{1\times T^m}$ denotes a row vector of all ones in length $T^m$.
We estimate the cosine similarity 
$s^m\!=\!(\frac{1}{T^m}\sum_{t'}\mathbf{r}^m_{t'}\cdot \frac{1}{n_I+t}\sum_{t'}\mathbf{r}_{t'})/(\Vert\frac{1}{T^m}\sum_{t'}\mathbf{r}^m_{t'}\Vert\Vert\frac{1}{n_I+t}\sum_{t'}\mathbf{r}_{t'}\Vert)$,
where a dataset is described as 
the mean of embeddings of all observations. 
Apart from liberating practitioners from manually defining meta-features of a dataset, the mean 
is even more descriptive and pertinent to the HPO behaviours.

To conclude, the 
dataset-aware attention 
allows a target configuration to attend those similar observed configurations of related datasets.
Though the dataset-aware attention raises the time complexity to $\mathcal{O}(n_\mathcal{X}(n_I\!+\!t\!+\!\sum_{m=1}^MT^m))$, the training can be approximately linear by conducting the attention in parallel.
Moreover, the TNP 
optimizes 
hyperparamters 
in significantly less 
SMBO
trials, namely a smaller value of $t$.
\begin{algorithm}[t]
\SetAlgoLined
\SetKwInOut{Input}{Input}\SetKwInOut{Output}{Output}
\Input{Observations on $M$ datasets $\mathcal{H}^1_{T^1},\cdots,\mathcal{H}^M_{T^M}$; \# of trials $T$; acquisition function $a$; target configurations $\{\hat{\mathbf{x}}_{j}\}_{j=0}^{n_\mathcal{X}}$; meta update rate $\epsilon$; \# of initial configurations $n_I$.}
\Output{
Best hyperparameter configuration $\mathbf{x}^*$ found.}
  Randomly initialize $\tilde{\theta}$, $\{\tilde{\mathbf{x}}_{Ij}\}_{j=1}^{n_I}$, and set  $y^{*}\leftarrow 0$\;
  \For{$m=1,\cdots, M$}{
    Perform $k$ gradient steps on : $\theta^m_k=\tilde{\theta}-\alpha\nabla^k_{\theta}\mathcal{L}(\mathcal{H}^m_{T^m,h},\mathcal{H}^m_{T^m,\bar{h}}|\theta)$, 
    $\mathbf{x}^k_{Ij}\!=\!\tilde{\mathbf{x}}_{Ij}\!
-\!\alpha\nabla^k_{\mathbf{x}_{Ij}}\mathcal{L}_{I}(\{\mathbf{x}_{Ij}\}_{j=1}^{n_I}|\theta)$, $\forall j\!=\!0,\!\cdots\!,n_I$\;
    Update $\tilde{\theta}$ and $\{\tilde{\mathbf{x}}_{Ij}\}_{j=1}^{n_I}$: $\tilde{\theta}=\tilde{\theta}+\epsilon(\theta^m_k-\tilde{\theta})$, $\tilde{\mathbf{x}}_{Ij}=\tilde{\mathbf{x}}_{Ij}+\epsilon(\mathbf{x}^k_{Ij}
-\tilde{\mathbf{x}}_{Ij}
)
$\;
 }
 Query the values of $f$ at $\{\tilde{\mathbf{x}}_{Ij}\}_{j=1}^{n_I}$, and obtain the initial observation set $\mathcal{H}_0\!=\!\{(\tilde{\mathbf{x}}_{Ij}, \tilde{y}_{Ij})\}_{j=0}^{n_I}$\;
  \For{$t=1,\cdots,T$}{
     Fine-tune TNP by $k$ gradient steps: $\theta_k=\tilde{\theta}-\alpha\nabla^k_{\theta}\mathcal{L}(\mathcal{H}_{t-1,h},\mathcal{H}_{t-1,\bar{h}}|\theta)$\;
     Optimize the promising configuration using $\text{TNP}_{\theta_k}$:
     $\mathbf{x}_t \leftarrow \arg\max_{\mathbf{x}\in\{\hat{\mathbf{x}}_{j}\}_{j=0}^{n_\mathcal{X}}}a(\text{TNP}_{\theta_k}(\mathbf{x}))$\;
     Evaluate $y_t=f(\mathbf{x}_t)$ and update the observation set $\mathcal{H}_{t}\!=\!\mathcal{H}_{t-1}\!\cup\!\{(\mathbf{x}_{t},y_{t})\}$\;
     \If{$y_t>y^*$}{$\mathbf{x}^*,y^*\leftarrow \mathbf{x}_{t},y_{t}$\;}
     
  }
  return $\mathbf{x}^*$\;
 \caption{Transferable Neural Processes (TNP)} 
 \label{alg:tnp}
\end{algorithm}
\paragraph{Transferring parameters}
The TNP with dataset-aware attention, as introduced above, characterizes the similarity between observed configurations and a target configuration. Consequently, it learns an implicit and data-driven kernel that is analogous to the analytic and manually defined kernel in GPs, 
e.g., Mat\'ern-5/2. 
Again, learning the kernel at the beginning of SMBO is difficult, inasmuch as few training pairs $(\mathcal{H}_{t-1,h}, \mathcal{H}_{t-1,\bar{h}})$ is sampled from $\mathcal{H}_{t-1}$ with limited observations to optimize Eqn.~(\ref{eqn:loss}).
Sampling 
response functions from an underlying distribution $P$
by sampling different datasets
in Eqn.~(\ref{eqn:loss}) 
provides a remedy, in a manner that the parameters of the kernel are transferred from other datasets. 
Nonetheless, the globally shared kernel across datasets suffers from catastrophic forgetting 
as training proceeds, and 
runs counter to practical scenarios where 
historical datasets are likely from heterogeneous 
distributions. 

The hierarchical Bayesian model is qualified to alleviate the problems:
there is a global kernel $\tilde{\theta}$ on which 
each dataset-specific kernel $\theta^m$ for the $m$-th dataset is statistically dependent. 
It still empowers knowledge transfer across datasets while fitting a wide range of datasets with knowledge customization.  
Without loss of scalability and end-to-end training of neural networks, we follow the strategy of model agnostic meta-learning (MAML)~\cite{finn2017model} which has been proved its equivalence to hierarchical Bayesian inference~\cite{grant2018recasting}.
Specifically, 
a transferable initialization $\tilde{\theta}$ for parameters of TNP as
the global kernel is inferred, and the dataset-specific parameters $\theta^m_k$ as the customized kernel are further optimized (fine-tuned) 
in $k$ gradient steps, i.e.,
\begin{equation}
     \theta^m_k\!=\!\tilde{\theta}-\alpha\nabla^k_{\theta}\mathcal{L}(\mathcal{H}_{T^m,h}^m,\mathcal{H}^m_{T^m,\bar{h}}|\theta),  \phantom{a}
     \tilde{\theta}\!=\!\tilde{\theta}
+\epsilon(\theta^m_k-\tilde{\theta}).
\label{eqn:opt_theta}
\end{equation}
In 
the meta-training stage of Figure~\ref{fig:framework}, each time we sample the $m$-th dataset as the target 
and the rest of $M$ datasets as historical datasets. 
First, initialized with $\tilde{\theta}$, TNP 
optimizes $\mathcal{L}(\mathcal{H}^m_{T^m,h},\mathcal{H}^m_{T^m,\bar{h}}|\theta)$ in $k$ gradient steps, where we follow Section~\ref{sec:npm} by dividing the observation set $\mathcal{H}^{m}_{T^m}$ into two parts.
In turn, $\theta^m_k$ 
updates the 
initialization $\tilde{\theta}$. 
During meta-testing, it is straightforward to 
fine-tune $\tilde{\theta}$ 
on 
the observation set of the target dataset
$\mathcal{H}_{t-1}$, i.e., $\theta_k\!=\!\tilde{\theta}\!-\!\alpha\nabla^k_{\theta}\mathcal{L}(\mathcal{H}_{t-1,h},\mathcal{H}_{t-1,\bar{h}}|\theta)$, and then 
predict 
for 
target configurations using the TNP equipped with the parameters $\theta_k$.

\paragraph{Initializing SMBO with well-generalized configurations}
The initial configurations have been demonstrated crucial to the success of SMBO~\cite{lindauer2018warmstarting,wistuba2015learning} -- those initial configurations which achieve larger values of $f$ (here we discuss the maximization of $f$ in Eqn.~(\ref{eqn:hyperopt_1})) are prone to speed up the SMBO.
Fortunately, we are provided with 
$M$ 
observation sets $\mathcal{H}^1_{T^1},\!\cdots\!,\mathcal{H}^M_{T^M}$ which offer a treasure of the configurations with higher $f$ values. 
Taking the heterogeneity of datasets into account, 
we again
formulate the problem of learning initial configurations as a hierarchical Bayesian inference problem.
Similar to inferring $\tilde{\theta}$ in Eqn.~(\ref{eqn:opt_theta}), we learn 
a 
set of well-generalized initial configurations $\{\tilde{\mathbf{x}}_{Ij}\}_{j=1}^{n_I}$ which are fine-tuned for each 
dataset.
The only difference is the loss with regards to
initial configurations 
$\{{\mathbf{x}}_{Ij}\}_{j=1}^{n_I}$, 
which 
enforces 
the predictions of at least one of the initial configurations 
to be 
maximized, i.e., 
$
\mathcal{L}_{I}(\{\mathbf{x}_{Ij}\}_{j=1}^{n_I}|\theta) = \sum_{j=1}^{n_I}\frac{e^{\alpha  {\mu}_{Ij}}}{\sum_{j'}^{n_I}e^{\alpha  {\mu}_{Ij'}}} {\mu}_{Ij}.
$
The reason why we impose the softmax 
is to not only maximize the $f$ value  but also preserve the diversity of the initial 
configurations, so that TNP as the surrogate can be well trained with comprehensive coverage to approximate the response function $f$. 
The overall learning algorithm is presented in Algorithm~\ref{alg:tnp}.

\section{Experiments}
\subsection{Experimental Setup}
\textbf{Datasets} 
We evaluate TNP on two categories of datasets, i.e., OpenML and computer vision datasets. 
First of all, we consider the OpenML~\cite{vanschoren2014openml} platform which contains a large number of datasets covering a wide range of applications.
Due to time constraints, we select 100 supervised classification datasets that have fewer than 100,000 instances and no missing values.
The training, validation, and test sets of each dataset are exactly the same as OpenML provides.
The hyperparameters are optimized on validation sets, while we compare different HPO methods by reporting the performance of  the best configuration returned on test sets.
For comparison with those baselines using meta-features to measure the similarity between datasets, we extract a list of meta-features for each dataset following the Table 1 in~\cite{wistuba2015learning}.
We aim to improve the classification accuracy of Logistic Regression (LR)~\cite{lecun1998gradient} on all OpenML datasets. 
The dimension of the hyperparameter space is four, including the learning rate $\eta\in[2^{-6},2^0]$ for SGD, the l2-regularization coefficient $r_2\in[0,1]$, the batch size $B\in[20, 2000]$, and the dropout ratio $\gamma\in[0,0.75]$.

\begin{figure}[t]
	\centering
	\begin{subfigure}[c]{0.47\textwidth}
		\centering
		\includegraphics[scale=0.5]{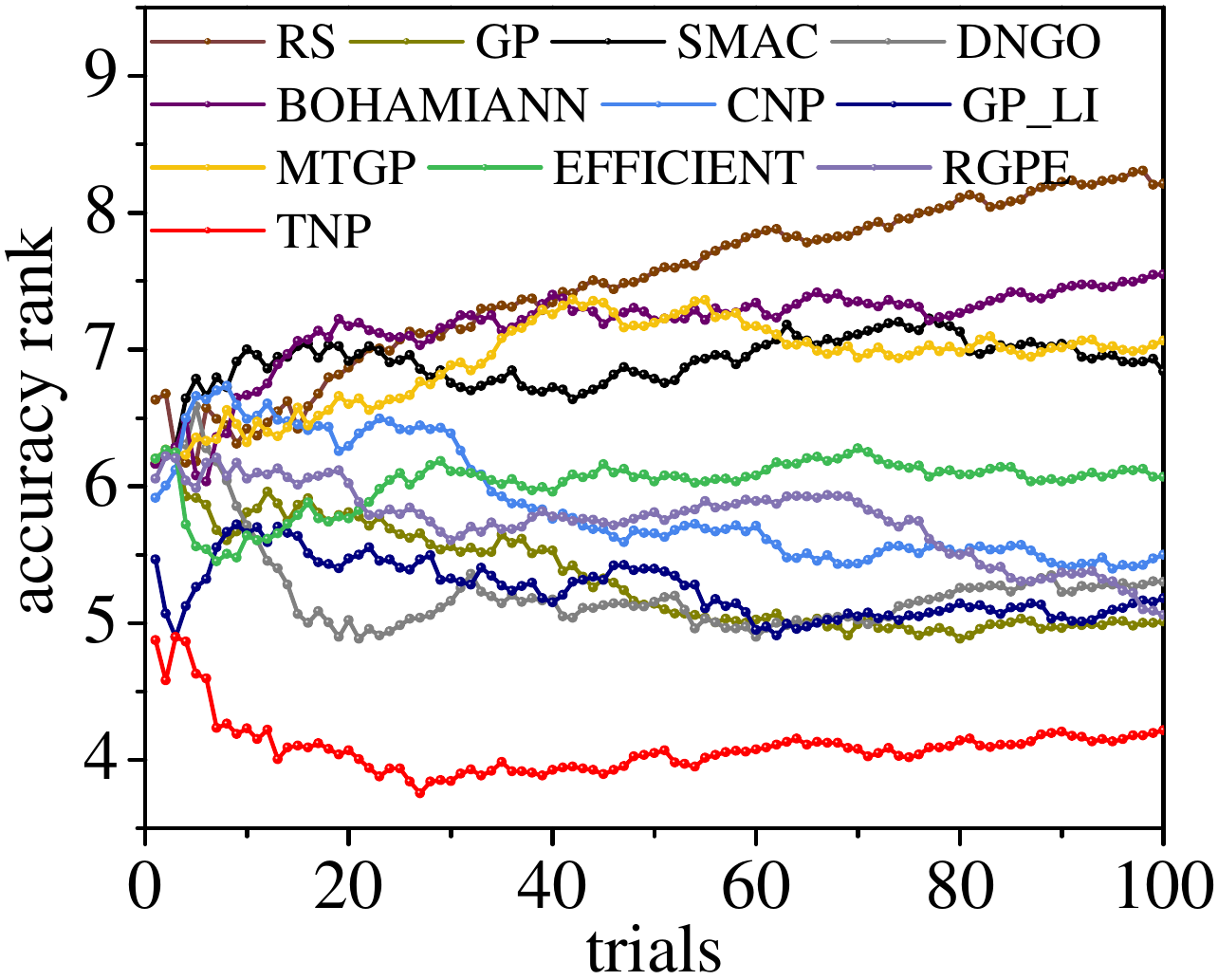}
		\caption{\label{fig:rank}: Average rank}
	\end{subfigure}
	\begin{subfigure}[c]{0.47\textwidth}
		\centering
		\includegraphics[scale=0.5]{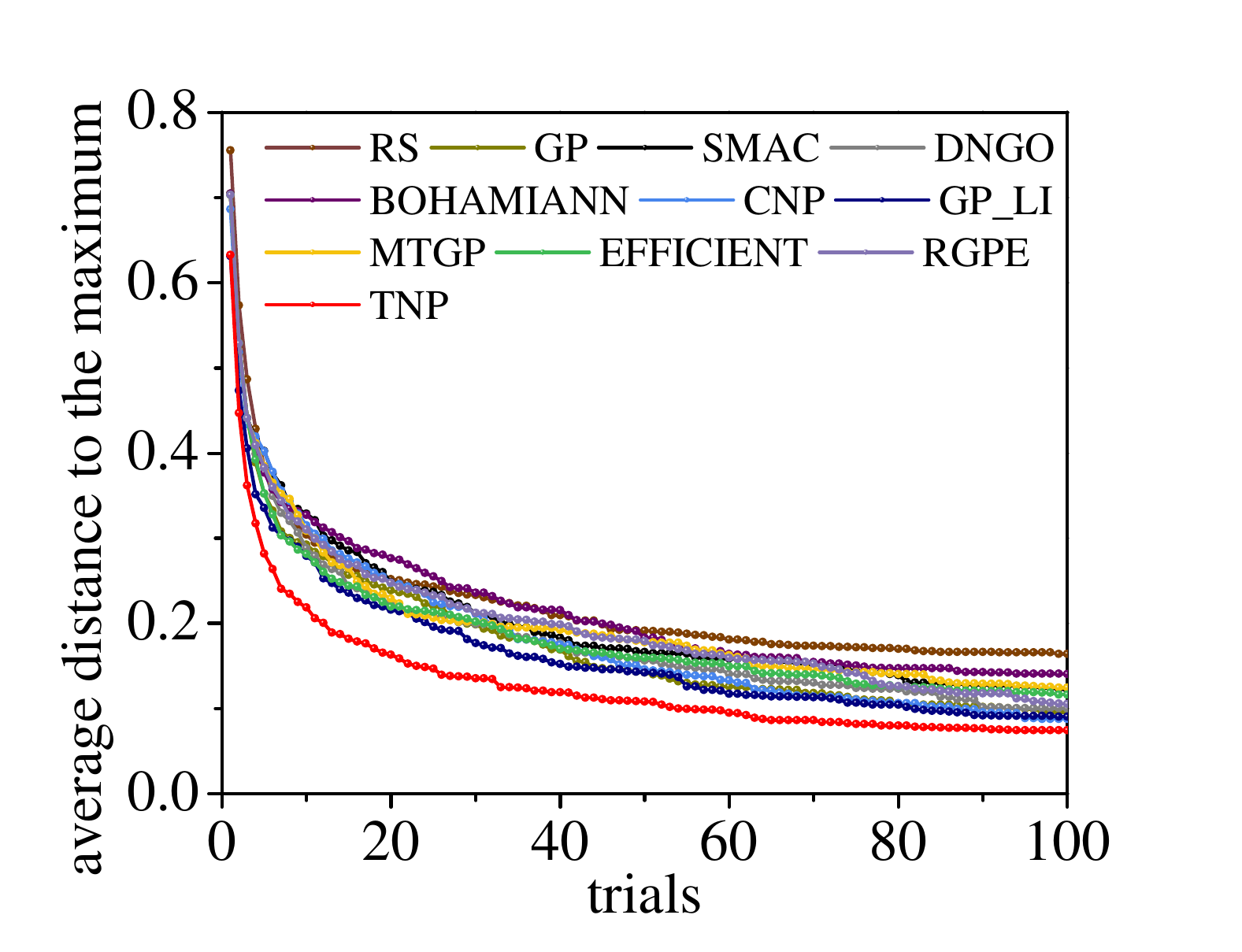}
		\caption{\label{fig:adtm}: Avereage distance to the maximum}
	\end{subfigure}
	\caption{Average ranks and distances to the maximum over 100 OpenML datasets across $100$ trials.}
	\label{fig:rank_adtm}
\end{figure}

Besides OpenML, we also investigate the effectiveness of TNP on three popular computer vision datasets, including CIFAR-10~\cite{krizhevsky2009learning}, MNIST~\cite{lecun1995learning}, and SVHN~\cite{netzer2011reading}. 
We take the last 10,000, 10,000, and 6,000 training instances as the validation set
for CIFAR-10, MNIST, and SVHN, respectively.
Each of them is also described with meta-features, which is the same as an OpenML dataset.
Here we focus on a three layer convolutional neural network in which each layer consists of a convolution with batch normalization and ReLU activation functions followed by max pooling. All convolutions have the filter size of $5\times 5$.
We tune five hyperparameters including the learning rate $\eta\in[2^{-6},2^0]$ for Adam, the batch size $B\in[32, 512]$, and the number of hidden units for the three layers, $d_1,d_2,d_3\in[2^4,2^8]$, respectively. 

\textbf{Baselines}
\begin{figure*}[b]
	\centering
	\begin{subfigure}[c]{0.325\textwidth}
		\centering
		\includegraphics[scale=0.37]{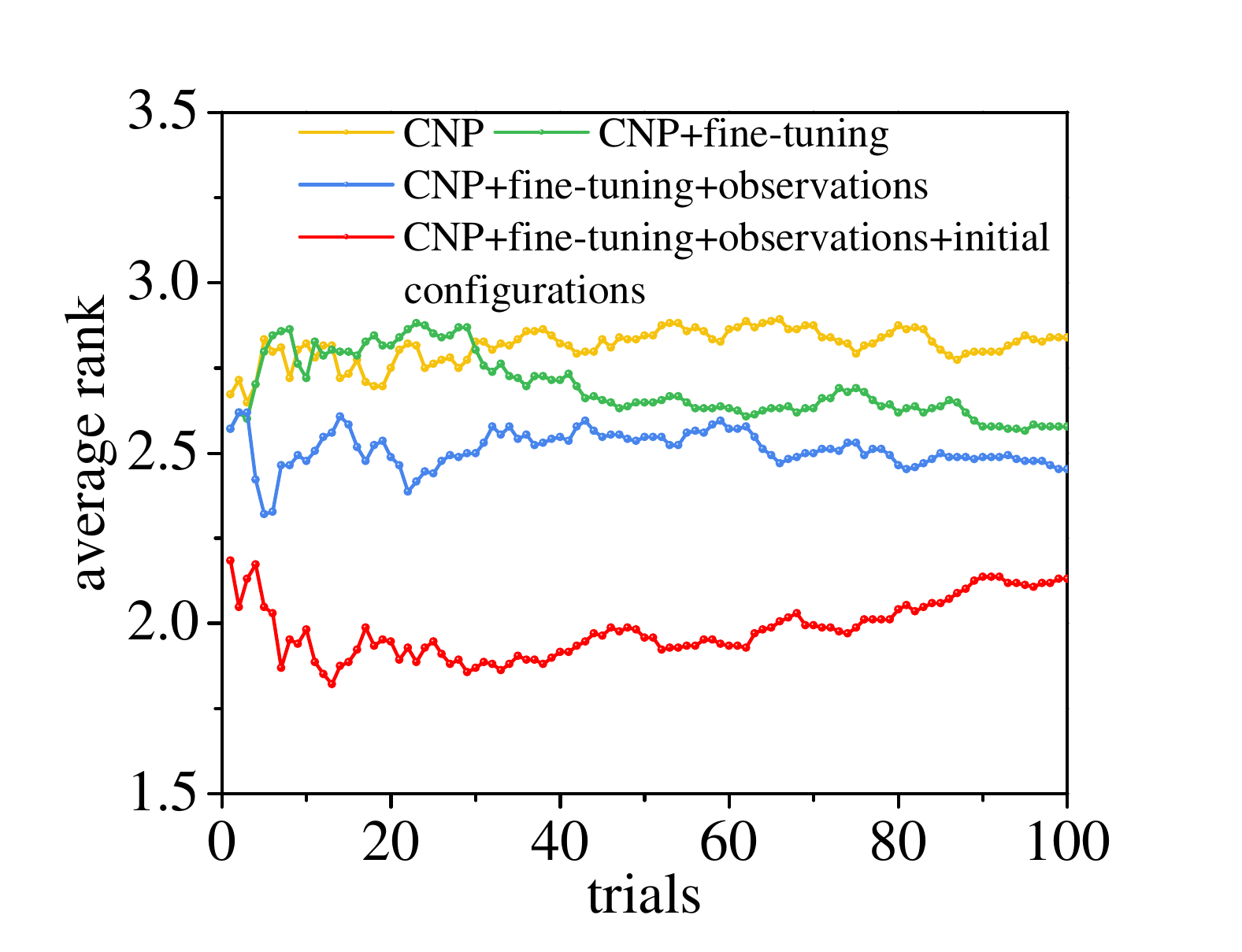}
		\caption{\label{fig:ablation}: Varying different components}
	\end{subfigure}
	\begin{subfigure}[c]{0.325\textwidth}
		\centering
		\includegraphics[scale=0.37]{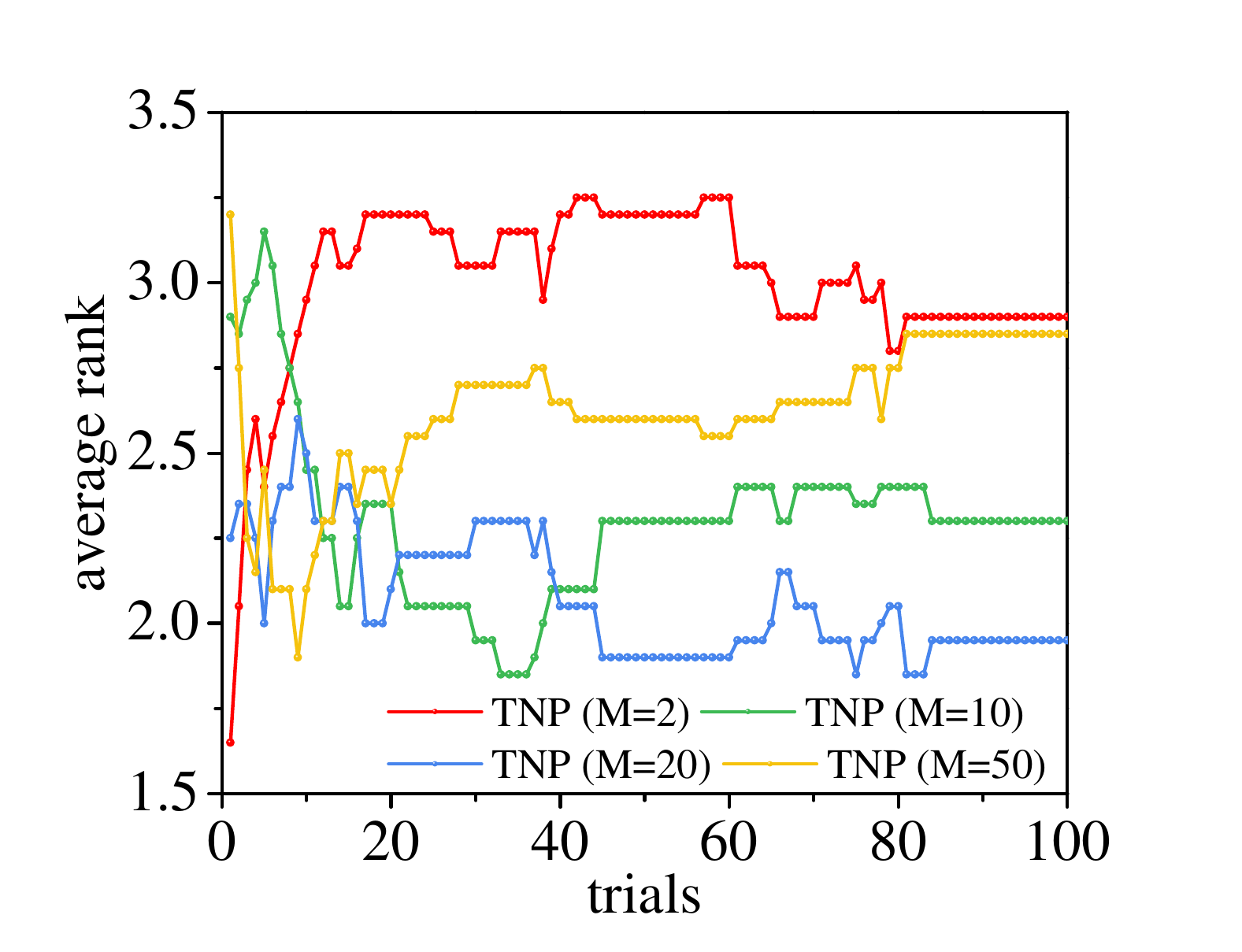}
		\caption{\label{fig:meta_num}: Varying the \# of meta-datasets}
	\end{subfigure}
	\begin{subfigure}[c]{0.325\textwidth}
		\centering
		\includegraphics[scale=0.37]{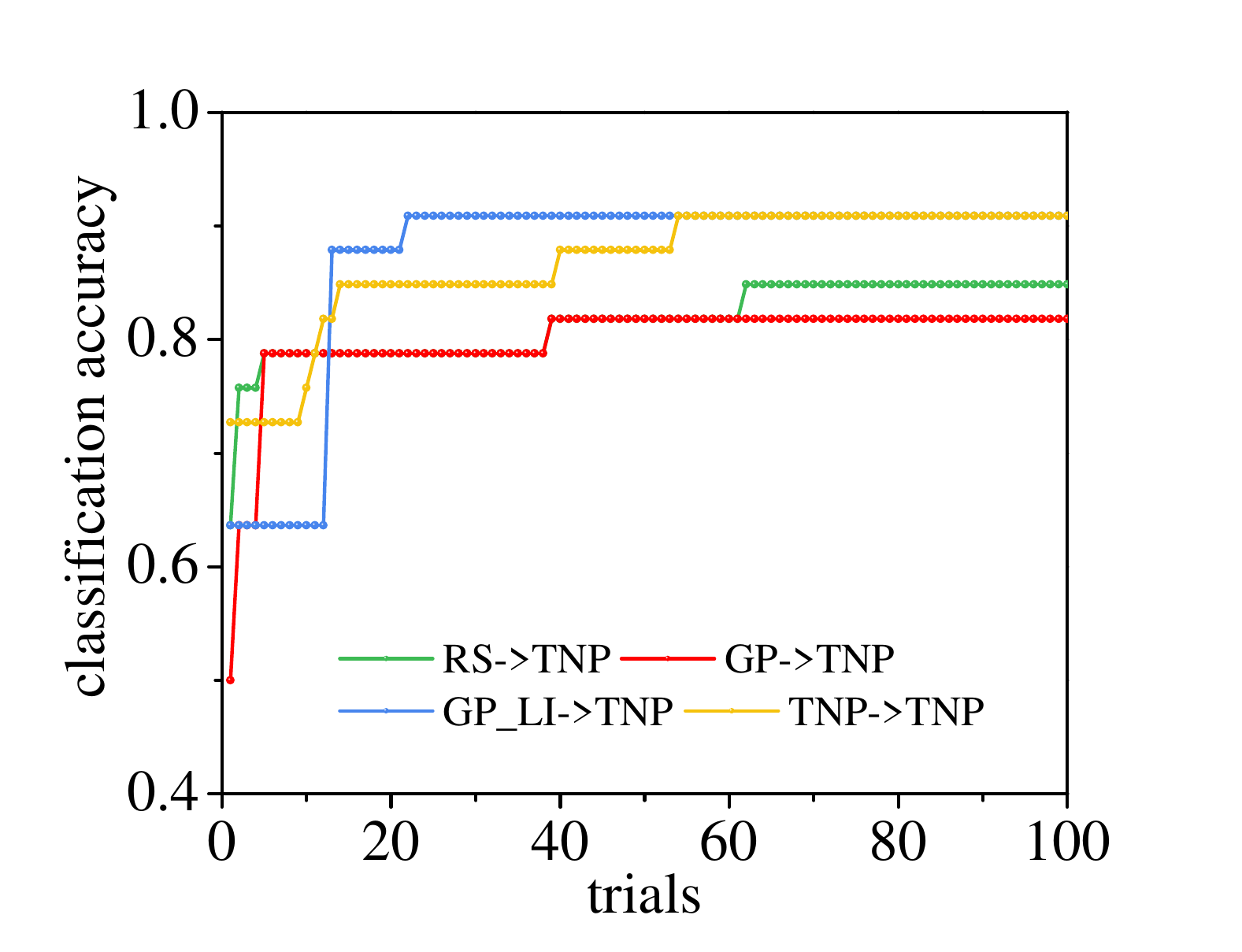}
		\caption{\label{fig:observation}: Varying the base method}
	\end{subfigure}
	\caption{Varying different components, t he number of meta-datasets, and the base method in TNP.}
	\label{fig:illustration}
\end{figure*}
We consider 11 baseline methods for comparison. 
Note that all of these methods including ours are based on the SMBO framework and use the expected improvement (EI) as the acquisition function. 
We categorize the baselines into four groups based on the surrogate model and whether knowledge is leveraged from other datasets. 
1) \emph{No surrogates}: random search~\cite{bergstra2012random}(\textbf{RS});
2) \emph{Surrogates without neural networks}:
Gaussian Processes~\cite{snoek2012practical} with a Mat\'ern-5/2 kernel (\textbf{GP}) and random forests~\cite{hutter2011sequential} (\textbf{SMAC});
3) \emph{Surrogates with neural networks}:
\textbf{DNGO}~\cite{snoek2015scalable} and   \textbf{BOHAMIANN}~\cite{springenberg2016bayesian} with recommended parameters;
\begin{figure*}[h]
	\centering
	\begin{subfigure}[t]{0.32\textwidth}
		\centering
		\includegraphics[scale=0.37]{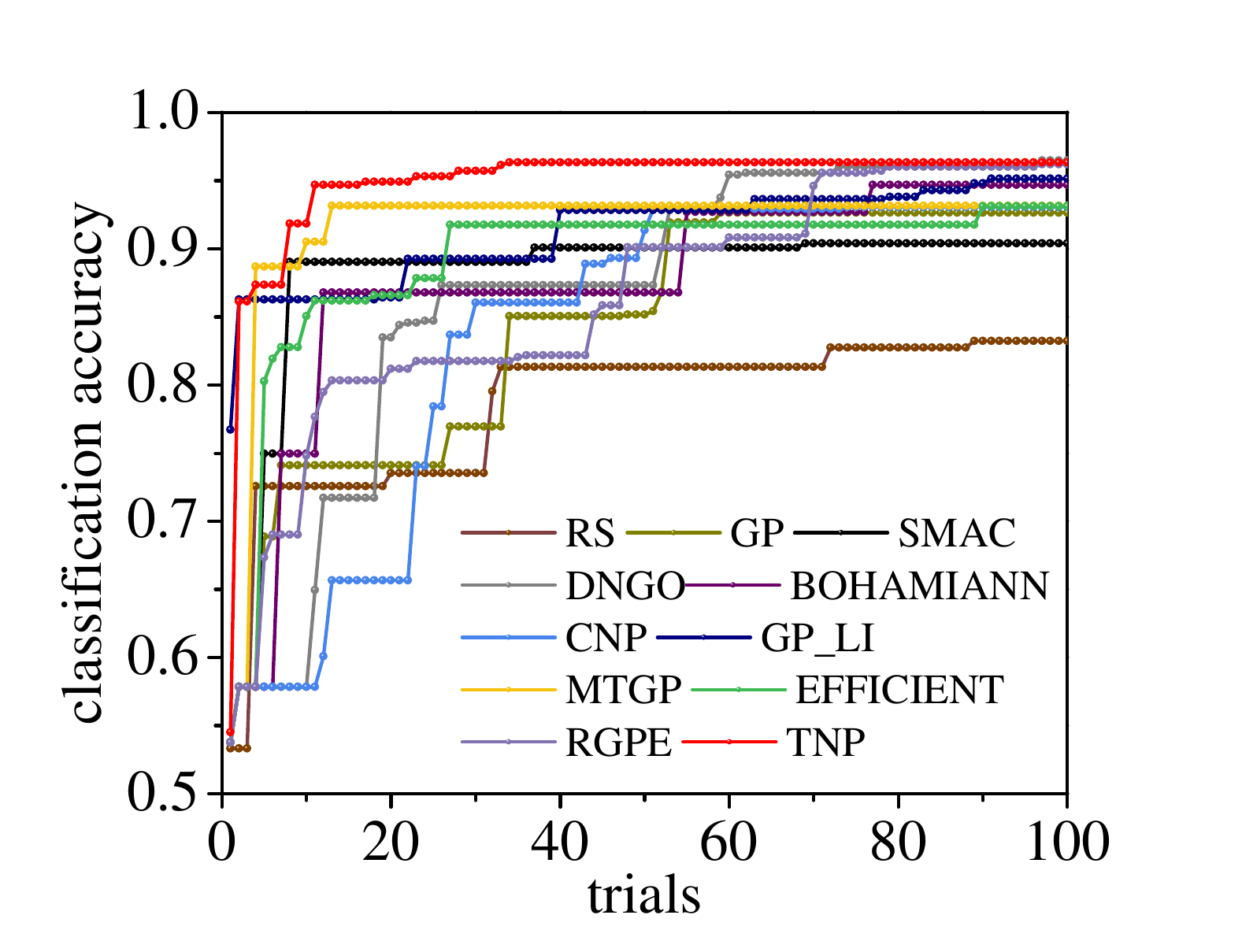}
		\caption{\label{fig:wine}: wine+oh5.wc$\rightarrow$ kr-vs-kp}
	\end{subfigure}
	\begin{subfigure}[t]{0.32\textwidth}
		\centering
		\includegraphics[scale=0.37]{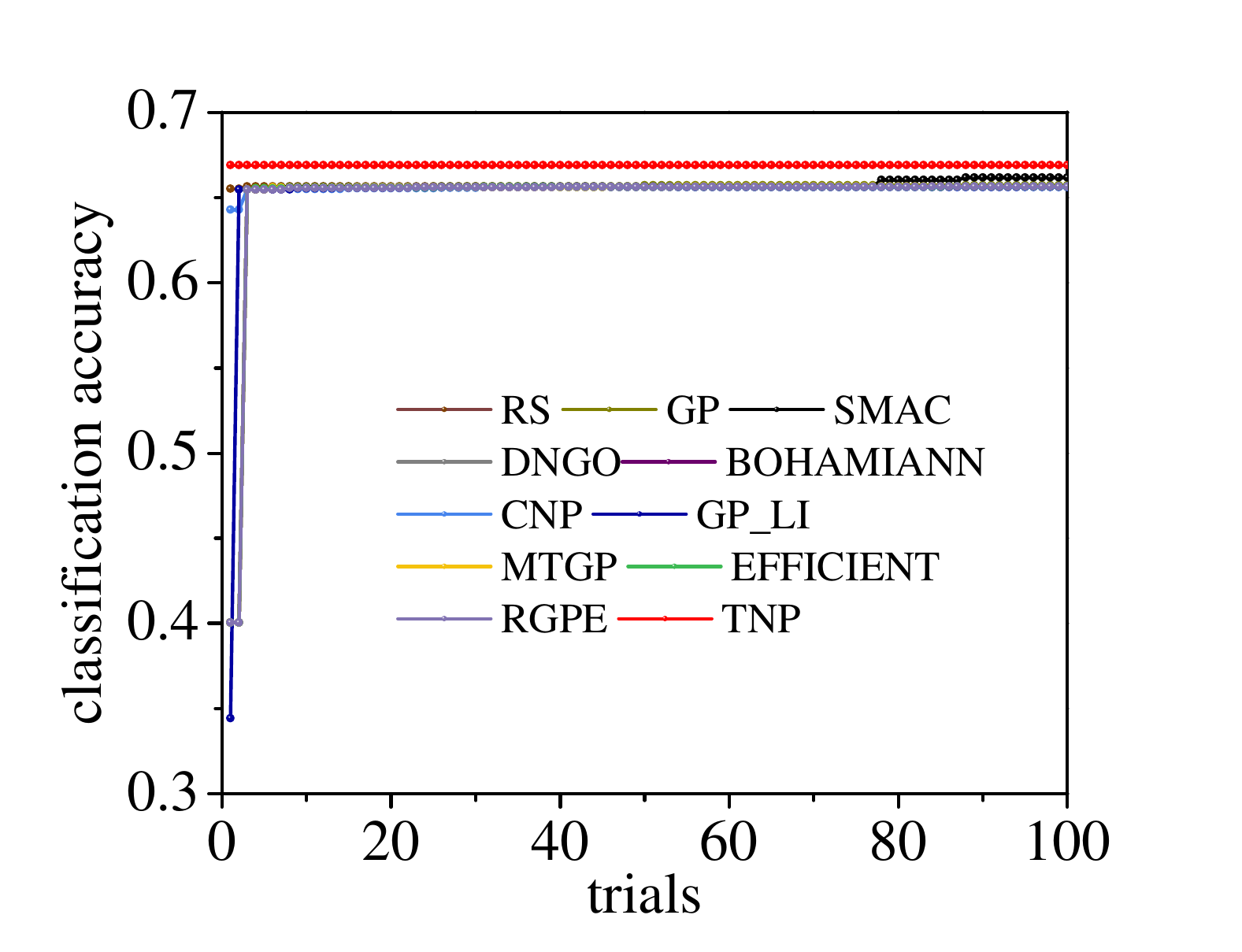}
		\caption{\label{fig:bng}: BNG(breast-w)+mfeat-zernike  $\rightarrow$BNG(tic-tac-toe)}
	\end{subfigure}
	\begin{subfigure}[t]{0.32\textwidth}
		\centering
		\includegraphics[scale=0.37]{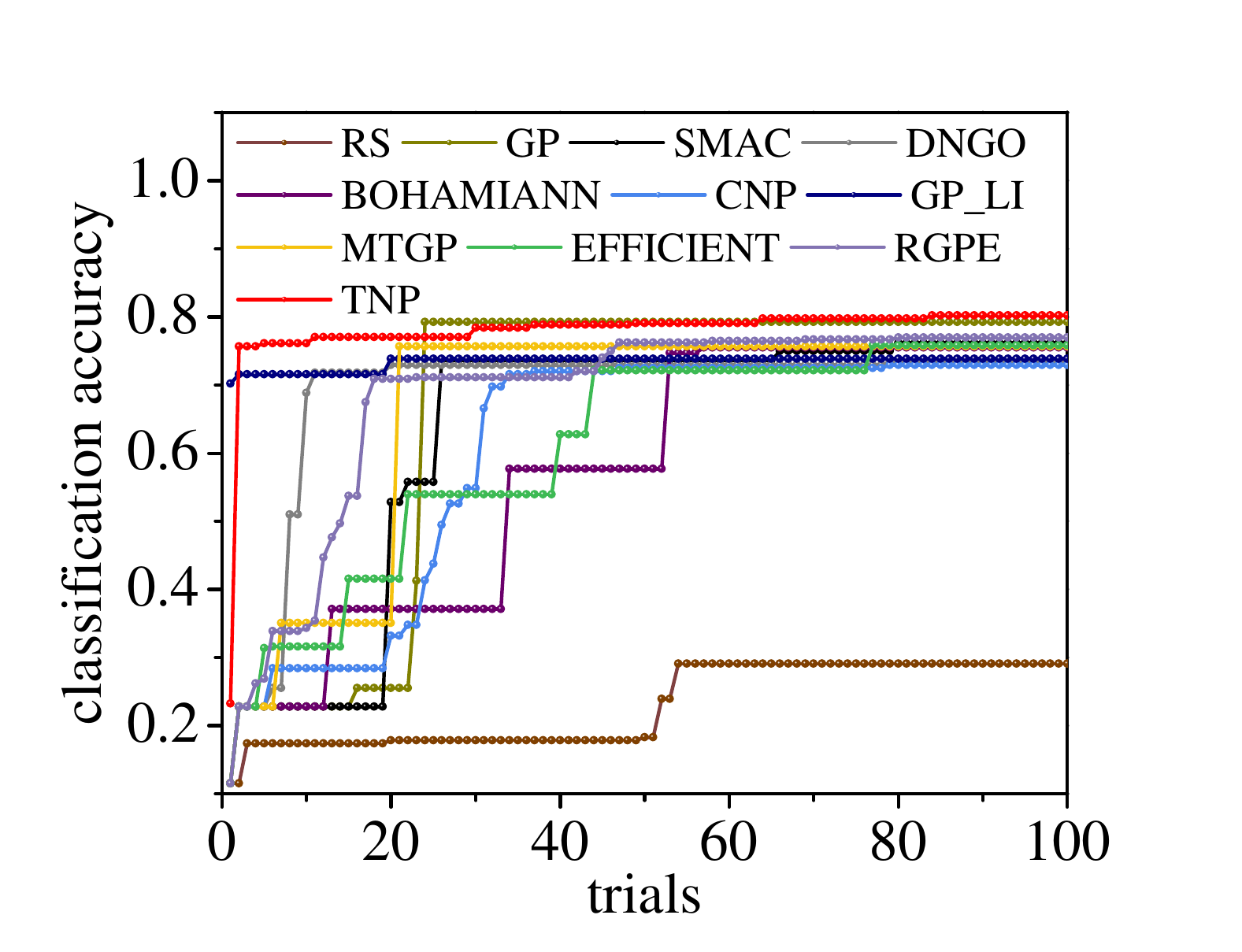}
		\caption{\label{fig:mfeat}: \text{fri}\_c4\_100+\text{fri}\_c4\_25  $\rightarrow$  mfeat-fourier}
	\end{subfigure}
	\caption{Comparison of the maximum accuracies achieved so far on three randomly selected datasets.}
	\label{fig:examples}
\end{figure*}
4) \emph{Surrogates with knowledge transfer}:
multitask GPs~\cite{swersky2013multi} (\textbf{MTGP}) and \textbf{EFFICIENT}~\cite{yogatama2014efficient} that transfer observations without and with meta-features, respectively,  
ranking-weighted Gaussian Process
ensemble~\cite{feurer2018scalable} (\textbf{RGPE}) that transfers parameters from past GPs, conditional neural processes~\cite{garnelo2018conditional} (\textbf{CNP}) with the same structure and parameter configuration with ours but a globally shared kernel, 
and 
\textbf{GP\_LI}~\cite{wistuba2015learning} that leverages past observations to learn initial configurations.

\textbf{Evaluation Metrics}
We compare 
in terms of the maximum classification accuracy achieved so far, the average rank over all datasets indicating the rank of a method, and the scaled average distance to the maximum across iterations. For details of the last two metrics, please refer to~\cite{bardenet2013collaborative} and~\cite{wistuba2015learning}, respectively.
The lower (smaller) the rank (distance) is, the more effective a method is.

\textbf{Network Setup}
The encoder, the decoder, and the attention embedding function $g$ are all implemented as a two layer multilayer perceptron with $[128, 128]$ hidden units, which indicates $r=128$. 
Following~\cite{garnelo2018conditional,kim2019attentive}, we first pre-train the networks by sampling $30,000$ batches of $n_\mathcal{X}$ dimensional GP functions with the length scale $l\sim U[0.3, 1.0]$ and the kernel scale $\sigma=1.0$.
Note that we set the batch size, the number of gradient steps $k$, and the learning rate $\alpha$ for Adam, and the meta update rate $\epsilon$ to be 64, 10, 1e-5, and 0.01, respectively.

\subsection{Results on OpenML Datasets}
\textbf{Effectiveness of hyperparameter optimization}
For each of the OpenML datasets, we obtain its history set of observations by running GPs to optimize the hyperparameters of LR on it within $100$ trials.
Taking
each
of the $100$ datasets as the target, 
we first randomly sample $M\!=\!2$ of the 99 others as historical meta-datasets, and leverage the 
two 
history sets to improve the effectiveness of ours as well as other transfer learning baselines.
Figure~\ref{fig:rank_adtm} shows the average rank and average distance to the maximum over all datasets. 
In general, we can see that the proposed TNP consistently and significantly outperforms other baselines, especially over a wide range of OpenML datasets.
Unsurprisingly, random search without a surrogate model performs the worst, even if we observe that it 
occasionally 
stands out 
on
a specific dataset, e.g., BNG (tic-tac-toe) as shown in Figure~\ref{fig:bng}.
GPs proves itself almost the most robust algorithm without knowledge transfer, as long as a sufficient number of observations have been collected. 
Consequently, despite the superioty of some baselines at the beginning ($<40$), e.g., DNGO, GPs becomes increasingly powerful.
Since none of the transfer learning baselines simultaneously transfers parameters, observations and initial configurations, they seem to be competent only at the very beginning. 
Except GP\_LI and TNP, other baselines share the same set of initial configurations. 
Though GP\_LI approaches TNP within $n_I\!=\!3$ initial configurations, which demonstrates that it is capable of learning 
competitive 
historical 
datasets and well-generalized parameter initializations into consideration.

We also randomly select three datasets and compare the maximum classification accuracies achieved so far by different algorithms in Figure~\ref{fig:examples}.
Notice that the performance of all baselines varies from dataset to dataset.
MTGP surprisingly outperforms all the baseline methods in Figure~\ref{fig:wine}, but all baselines are comparable in Figure~\ref{fig:bng}.
The datasets themselves explain such difference: kr-vs-kp with 3,196 instances in a dimension of 37 is more challenging than BNG (tic-tac-toe) with 39,366 instances of only 10 features, so that an optimal hyperparameter can be easily acquired for BNG (tic-tac-toe). 
Even in this case, TNP quickly learns a remarkable initial configuration, by leveraging observations in BNG (breast-w) and mfeat-zernike. 
In particular, we analyze the similarity vector between BNG (tic-tac-toe) and the other two datasets, i.e., $\mathbf{s}$ defined in Section~\ref{section:kt}, whose values are $0.5439$ and $0.1174$ in terms of BNG (breast-w) and mfeat-zernike, respectively.
By precisely modelling the similarity vector, TNP discovers more transferable knowledge between similar datasets but simultaneously alleviates negative transfer between wildly dissimilar ones.
\begin{figure*}[t]
	\centering
	\begin{subfigure}[c]{0.32\textwidth}
		\centering
		\includegraphics[scale=0.37]{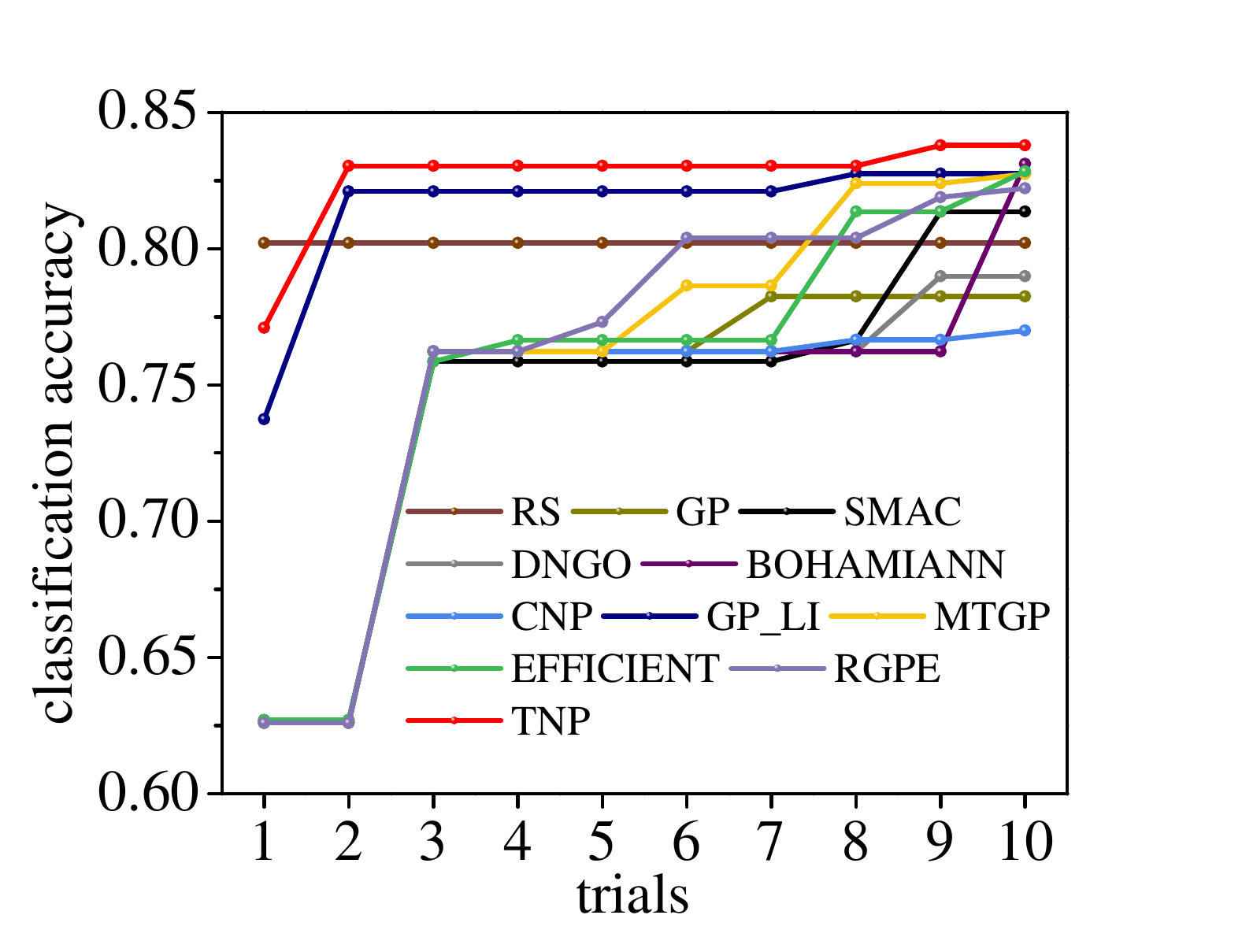}
		\caption{\label{fig:collaboration_shot}: CIFAR-10}
	\end{subfigure}
	\begin{subfigure}[c]{0.32\textwidth}
		\centering
		\includegraphics[scale=0.37]{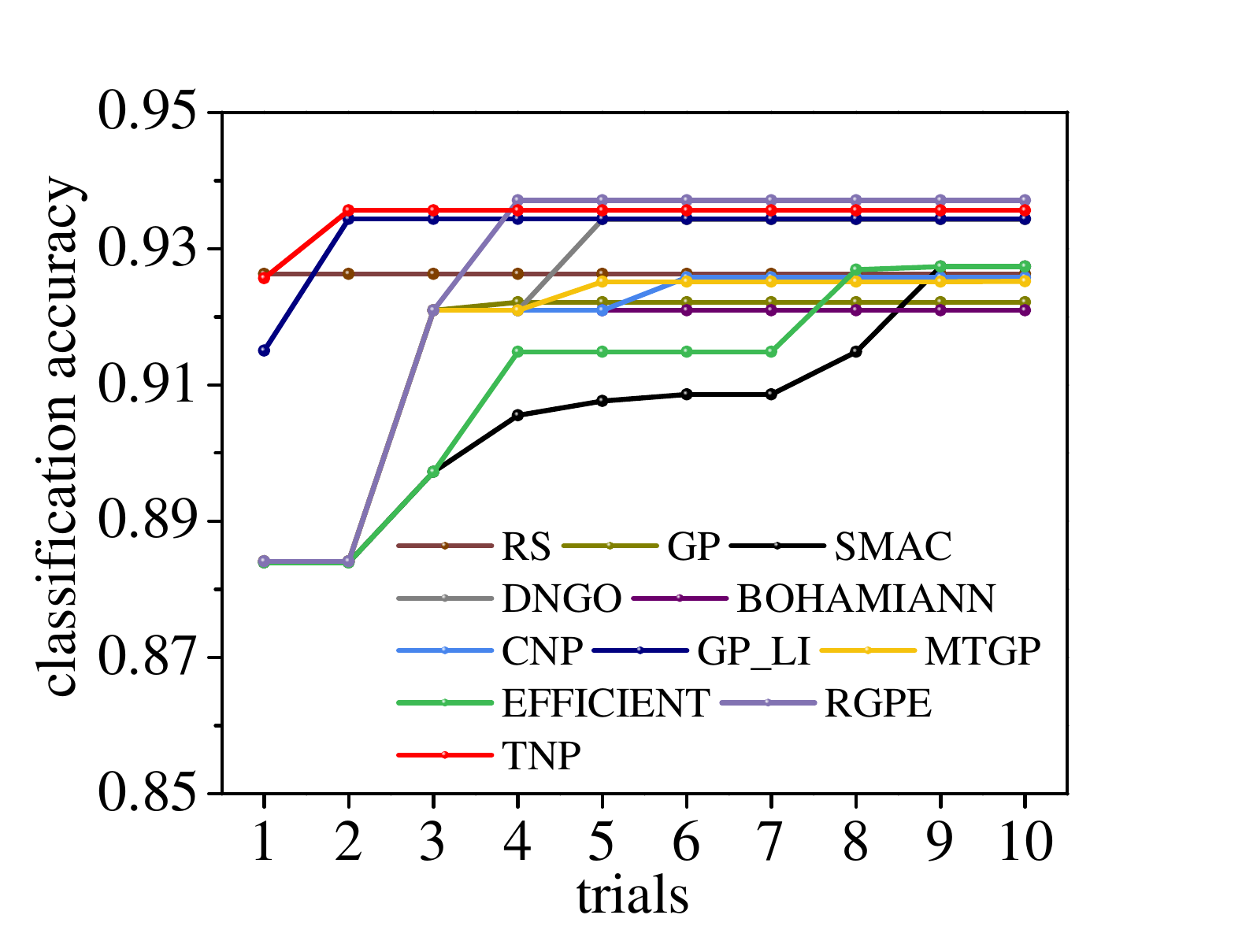}
		\caption{\label{fig:reddit_shot}: SVHN}
	\end{subfigure}
	\begin{subfigure}[c]{0.32\textwidth}
		\centering
		\includegraphics[scale=0.37]{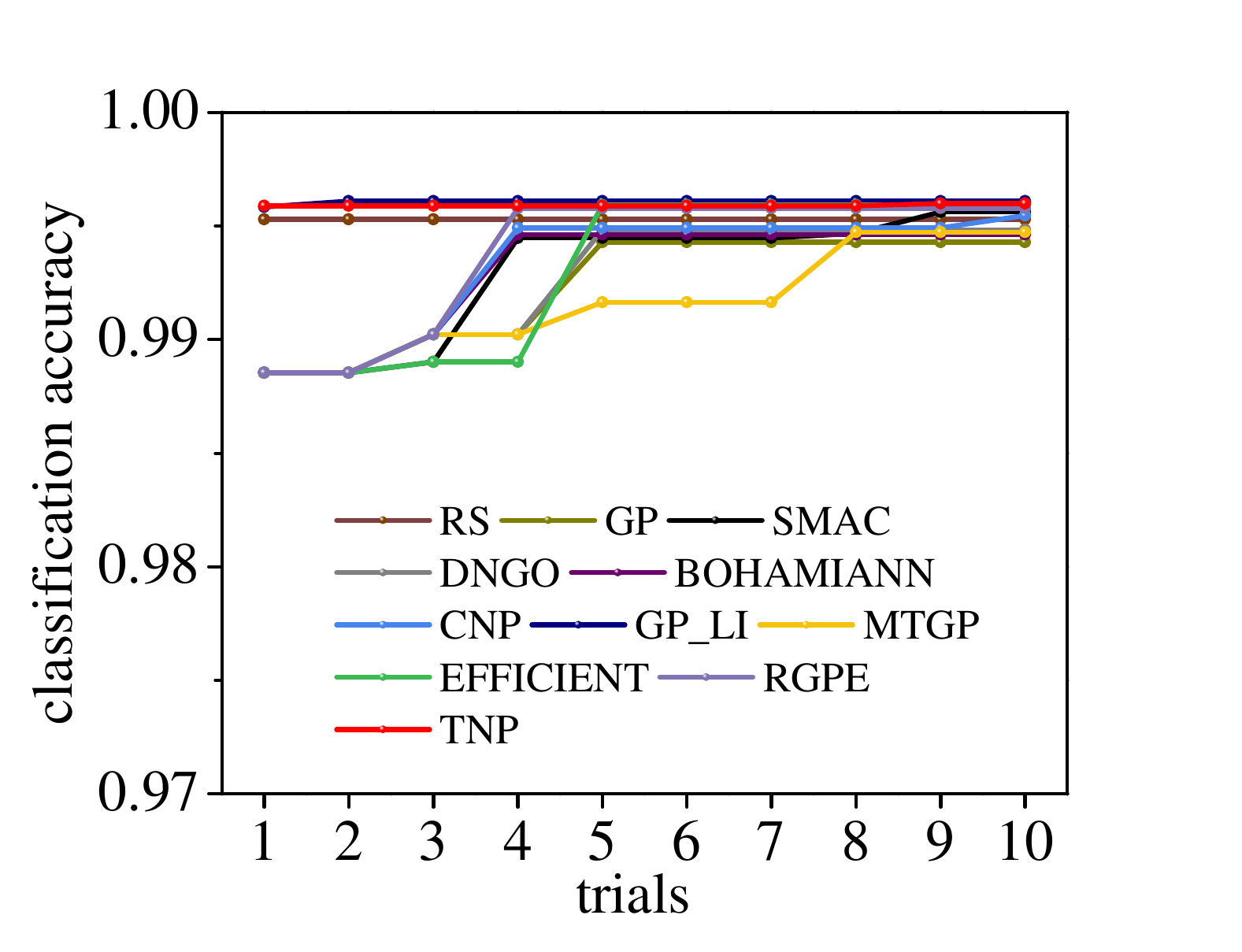}
		\caption{\label{fig:reddit_shot}: MNIST}
	\end{subfigure}
	\caption{Comparison of the maximum accuracies achieved so far on three computer vision datasets.}
	\label{fig:illustration}
\end{figure*}

\textbf{Ablation Studies}
First, we aim to study the influence of different components on 
the performance of TNP.
As shown in Figure~\ref{fig:ablation}, directly applying CNP 
with a globally-shared kernel 
does not guarantee efficient HPO. 
Incorporating the strategy of MAML for fine-tuning the kernel to be dataset-specific obviously mitigates the problem.
Most importantly, leveraging previous observations via the dataset-aware cross attention 
and learning a set of well-generalized initial configurations substantially boost the effectiveness of TNP.
Second, we vary the number of historical datasets, i.e., $M$.
Figure~\ref{fig:meta_num} tells that more datasets generally contribute more to improve the HPO. 
However, if the number of historical datasets is too large ($M=50$), 
the performance degrades a little, which is possibly due to more noisy 
observations from other datasets.
As mentioned above, the history set of each dataset is obtained by running GPs on it.
Here we are motivated to study how the base method used to produce the history set, e.g., GPs here, influences the performance of TNP.
Take the ``lymph'' dataset as an example.
The results in Figure~\ref{fig:observation} further guard the effectiveness of TNP regardless of the base method. 
Interestingly, the TNP can leverage knowledge from either the good or the bad (RS); the TNP
dependent on the history sets produced by a more effective base method, say TNP itself, is prone to outperform, provided with more insightful observations by the superior base method.

\subsection{Results on Computer Vision Datasets}
Considering the computational cost of each trial, we evaluate the accuracies of all baselines within only 10 trials, being more practical in real-world deployment. 
Regarding each of the three as the target,  the other two datasets act as $M=2$ historical datasets.
As the most challenging dataset, CIFAR-10 benefits the most from the other two, which is evidenced by the superior performance of both GP\_LI and TNP. 
Although RS occasionally and fortunately obtains a sub-optimal configuration, RS without a surrogate model gets stuck in the local optimum and cannot achieve higher performance.
An interesting observation is that most of transfer learning baselines perform better than GPs, while it is not the case in the HPO on OpenML datasets, which can be explained by the larger dimension of the hyperparameter space here.
Again TNP demonstrates its effectiveness on all of the three datasets - achieves competent classification accuracies within only 10 (and even the first three) trials. 
\section{Conclusion}
We introduced Transferable Neural Processes (TNP), a novel end-to-end hyperparameter optimization method which can leverage knowledge from past HPO observations on other datasets. 
With a dataset-aware attention unit, TNP can attentively borrow observations from those similar datasets, which prevents negative transfer. 
TNP, to the best of our knowledge, is the first to harness the collective power of transferring observations, learning a transferable initialization for the surrogate model, and initializing the SMBO with well-generalized configurations.
In particular, TNP enjoys the advantages of neural processes with high scalability.
In the future, we are committed to introduce the freeze-thaw mechanism into our model, so that the limited time budget can be allocated to those promising configurations which our model can quickly estimate via knowledge transfer.

\bibliography{ref}
\bibliographystyle{plain}

\end{document}